\documentclass{article}
\PassOptionsToPackage{numbers, compress}{natbib}

\usepackage[final]{neurips_2024}

\usepackage[utf8]{inputenc} 
\usepackage[T1]{fontenc}    
\usepackage{hyperref}       
\usepackage{url}            
\usepackage{booktabs}       
\usepackage{amsfonts}       
\usepackage{nicefrac}       
\usepackage{microtype}      
\usepackage{xcolor}         
\usepackage{graphicx} 
\usepackage{svg}
\usepackage{amsmath}

\title{Secure \& Personalized Music-to-Video Generation via CHARCHA}

\author{%
  Mehul Agarwal \\
  MLD, Carnegie Mellon University\\
  \texttt{mehul@cmu.edu} \\
  \And
  Gauri Agarwal \\
  MIT Media Lab \\
  \texttt{gauri\_al@mit.edu} \\
  \AND
  Santiago Benoit \\
  LTI, Carnegie Mellon University \\
  \texttt{sbenoit@cs.cmu.edu} \\
  \And
  Andy Lippman \\
  MIT Media Lab \\
  \texttt{lip@media.mit.edu} \\
  \And 
  Jean Oh \\
  RI, Carnegie Mellon University\\\texttt{hyaejino@andrew.cmu.edu} \\
}

\begin{document}

\maketitle

\begin{abstract}
Music is a deeply personal experience and our aim is to enhance this with a fully-automated pipeline for personalized music video generation. Our work allows listeners to not just be consumers but co-creators in the music video generation process by creating personalized, consistent and context-driven visuals based on lyrics, rhythm and emotion in the music. The pipeline combines multimodal translation and generation techniques and utilizes low-rank adaptation on listeners' images to create immersive music videos that reflect both the music and the individual. To ensure the ethical use of users' identity, we also introduce CHARCHA, a facial identity verification protocol that protects people against unauthorized use of their face while at the same time collecting authorized images from users for personalizing their videos. This paper thus provides a secure and innovative framework for creating deeply personalized music videos.
\end{abstract}

\begin{figure}[h!]
\centering
\includegraphics[width=\textwidth]{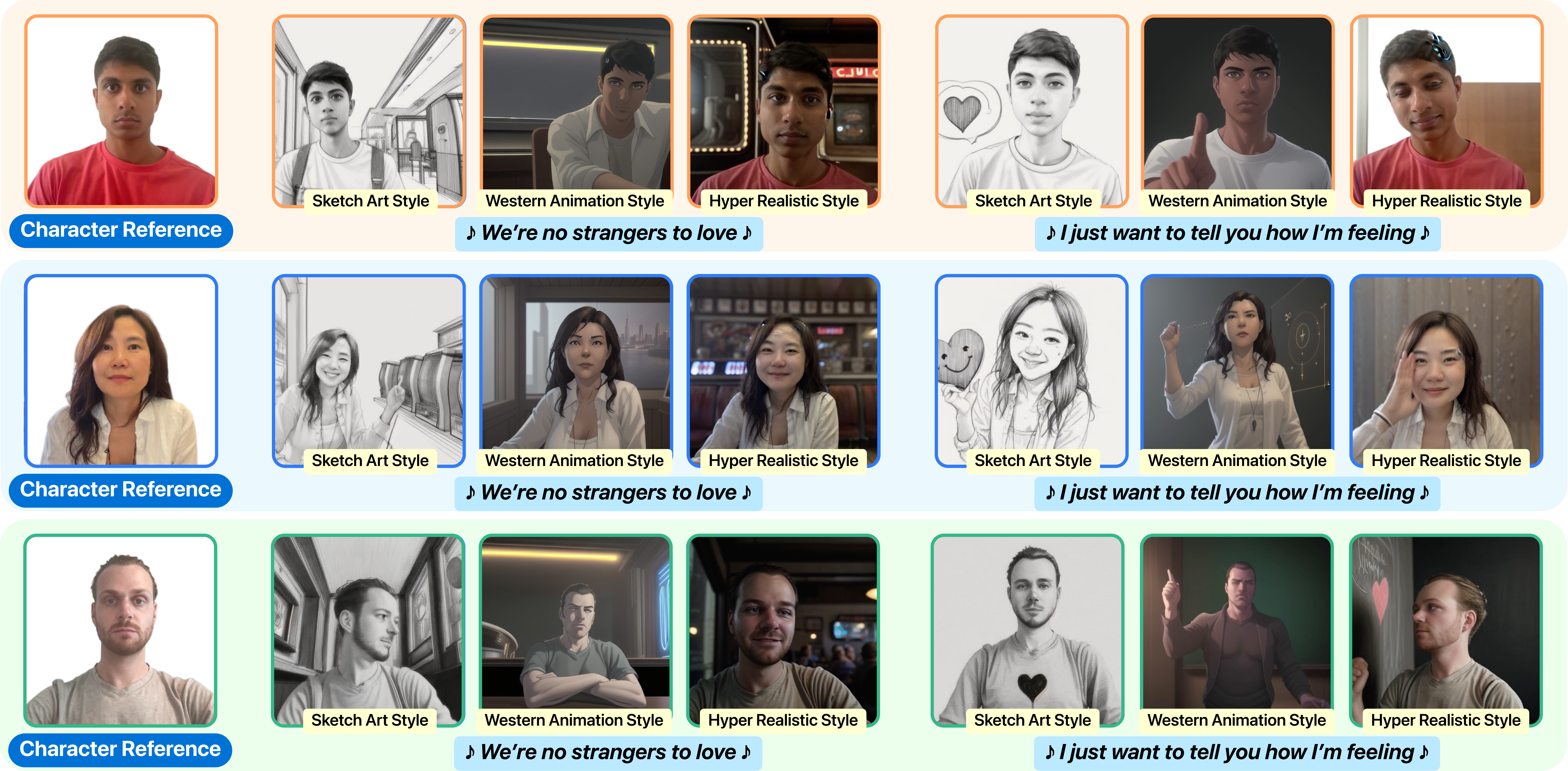}
\caption{Image stills and lyrics from generated music videos for Rick Astley's "Never Gonna Give You Up," with character reference from CHARCHA. The videos use Queratogray Sketch\cite{sketch}, Western Animation Diffusion\cite{west}, and Realistic Vision V5.1\cite{real_vis} checkpoint models}.
\label{personal_compare}
\end{figure}

\section{Introduction}
Our project enhances the music video creation process by utilizing the latest techniques in generative AI. We introduce a music video generation pipeline that combines multiple modalities-- audio, visual and language and efficiently generates personalized and secure music videos. We also introduce CHARCHA, an authentication protocol that builds on CAPTCHA's legacy. It is designed to be accessible to consenting humans yet challenging for machines and malicious actors. In turn, the data collected by CHARCHA is used to recreate digital avatars of the users. This approach offers a new way to experience music, allowing users to participate in creating videos while protecting against impersonation in the age of generative AI.

Traditionally, producing music videos requires significant resources, but our research extends multimodal diffusion-based models beyond text and image, aligning musical audio with visual synchronization. By exploring the temporal connections between these modalities, we aim to develop novel techniques and evaluation metrics that could have broader applications for multimodal video generation. Unlike previous works that have only explored the relationships between music and video; or text and video, our project takes a more integrated approach. 

With only the music audio file as input, MVP uses a zero-shot approach to extract the rhythm, melody, lyrics, and emotional context from the audio to generate visual content that is contextually aligned and synchronized with the music. Our fully automated pipeline integrates a range of pretrained models and APIs for tasks such as audio transcription, text-to-image diffusion, linear spherical interpolation, and music emotion recognition. Our framework is also designed to be versatile, accommodating various music genres and languages, and incorporating region-specific visual styles through diffusion-based models. Thus, our approach makes it easy for users, even with minimal technical expertise, to get started with generating their own music videos.

A key innovation in our approach is the inclusion of personalization, which allows listeners to become co-creators in the music video generation process. By using low-rank adaptation (LoRA) on the listener's images, we can incorporate their likeness into the videos, creating a more immersive and personalized experience. To facilitate this, we introduce CHARCHA, a facial identity verification protocol that serves a dual purpose: it verifies the user's identity by having them perform specific actions on camera, while simultaneously collecting these images to train the LoRA model. This approach not only safeguards individual privacy but also enhances personalization by directly integrating user-provided data. By combining these multimodal techniques, we aim to create a framework that not only advances the state of the art in music video generation but also prioritizes security and personalization.

\section{Related Works}
\textbf{Music to Video:} Music Video generation is currently done in closed source projects employing text-to-image models in the backend. Neural Frames's AI music generator \cite{neuralframes2023} uses Stable Diffusion, trained on 2.7 billion images, to generate video frames from text prompts. The platform employs stem extraction and audio-reactive features to synchronize visuals with music, allowing up to 10 visual parameters to be modulated by audio stems. Kaiber \cite{kaiber2023} also works similarly and offers more features like storyboard design for narrative flow and video transformation tools to easily modify visual styles. None of these closed source solutions, however, have automatic semantic understanding and take into account emotional information. Note that this work deviates from existing works transforming audio (in waveform or spectrogram) directly to the video data \cite{av1, av2, av3}, since we focus more on long form music, capturing emotion and narrative.

\textbf{Music Emotion Recognition (MER):} Delbouys et al. \cite{delbouys2018music} developed a deep neural network for music mood detection using audio spectrograms and lyric embeddings from 18,000 annotated tracks. They found mid-level fusion optimal for bimodal valence and arousal prediction. Our work employs a similar model, predicting valence and arousal from openSMILE features \cite{opensmile}. We subsitute the emotion extraction from lyric embeddings by directly feeding the lyrics to an LLM in image prompt generation.

\textbf{Text to Image and Video Models:} Latent Diffusion Models (LDMs) \cite{stablediff} use pretrained autoencoders' latent space to train diffusion models for image synthesis, reducing computational costs while maintaining quality. LDMs also achieve competitive performance in text-to-image tasks comparable to models like Google Imagen \cite{imagen} and OpenAI DALL-E 3 \cite{dalle}. Models such as Stable Diffusion XL \cite{SDXL} and Flux \cite{flux} extend this by allowing flexible stylization through finetuning and LoRA training. 

Text-to-video advancements have been driven by closed-source companies like Pika \cite{pika}, Luma Labs \cite{dreammachine}, RunwayML \cite{gen3alpha}, and OpenAI \cite{sora}, using diffusion transformer models \cite{dit}. Open-source alternatives like Stable Video Diffusion \cite{svd} and OpenSORA \cite{opensora} exist but have limitations in video length (~16 seconds) and quality \cite{opensora}. Thus for our purpose, we use Stable Diffusion 1.5 to generate image frames and apply spherical interpolation between frames to generate video.

\textbf{Authentication Protocols:} Authentication protocols like CAPTCHA \cite{captcha} present users with tasks that are challenging for machines but simple for humans. They have evolved from simple human-machine differentiation to sophisticated variants like reCAPTCHA \& hCAPTCHA \cite{captcha, recaptcha}. 

The advent of generative AI has shifted the paradigm, blurring the lines between human and machine-generated content. While generative AI offers unprecedented personalization opportunities, it also facilitates impersonation through DeepFakes and Celebrity LoRAs \cite{deepfake, celeblora}. Thus we introduce CHARCHA, a novel authentication method building on CAPTCHA's legacy to take advantage of the benefits of personalization while minimizing risk of identity manipulation and misuse. 

\section{Approach}

\subsection{Music to Video}

Given an audio file, we use a pretrained version of OpenAI's automatic speech recognition (ASR) model Whisper \cite{whisper} to obtain the lyrics at specific timestamps. We also pass the audio into a trained music emotion recognition model. 

Given the lyrics and the corresponding emotion at that timestamp, we use a pretrained text-to-image latent diffusion model (in our case, Stable Diffusion 1.5 \cite{stablediff}) to generate contiguous images based on the lyrics syncing them in a video. 

We condition our image prompts using Large Language Model conditioning, and additional negative prompts and style prompts. A few of these modeling considerations are discussed in more detail in the appendix.

\textbf{Emotion Extraction:} Using Music Emotion Recognition (MER), we analyze music's emotional content using the circumplex model of affect, which captures the ebb and flow of feeling through two key dimensions: arousal (the intensity of emotion) and valence (its positivity or negativity). This creates four emotional quadrants: Melancholy, Serene, Tense, and Euphoric \ref{emo}.
Our process (detailed in appendix):
\begin{enumerate}
    \item Extract openSMILE features from audio
    \item Use a neural network trained on the DEAM dataset \cite{DEAM} to predict arousal and valence
    \item Track emotional changes by monitoring position shifts in the arousal-valence space
    \item Combine emotional data with lyrics to generate image prompts via an LLM
\end{enumerate}

\textbf{LLM Conditioning:} We capture the timestamps with every change in lyric or emotion. We leverage ChatGPT 4o \cite{gpt4o} to transform song lyrics and rhythm sections into rich, nuanced image prompts. We feed the set of timestamps with the corresponding lyric and emotion to the LLM, prompting it to come up with story-based image prompts that preserve the narrative. This approach overcomes the limitations of literal lyric/emotion interpretation, creating visually evocative descriptions that capture the song's essence. For example, the lyric \textit{``I just wanna tell you how I'm feeling''} becomes
\textit{``The speaker stands beneath a night sky, gazing into their lover's eyes, conveying the overwhelming magnitude of their feelings.''} 

\begin{figure}[h!]
\centering
\includegraphics[width=0.8\textwidth]{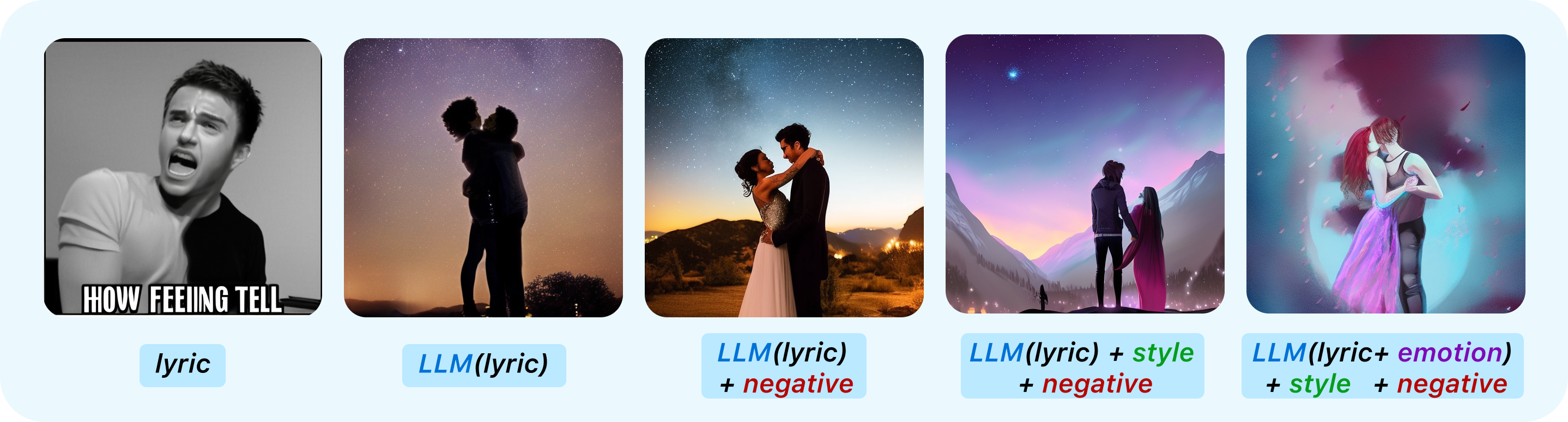}
\caption{Image generation based on the lyric "I just wanna tell you how I'm feeling", progressively incorporating LLM conditioning, negative prompting, style prompting, and emotion prompting}
\label{prompting}
\end{figure}

\begin{figure}[h!]
    \centering
\includegraphics[width=0.7\textwidth]{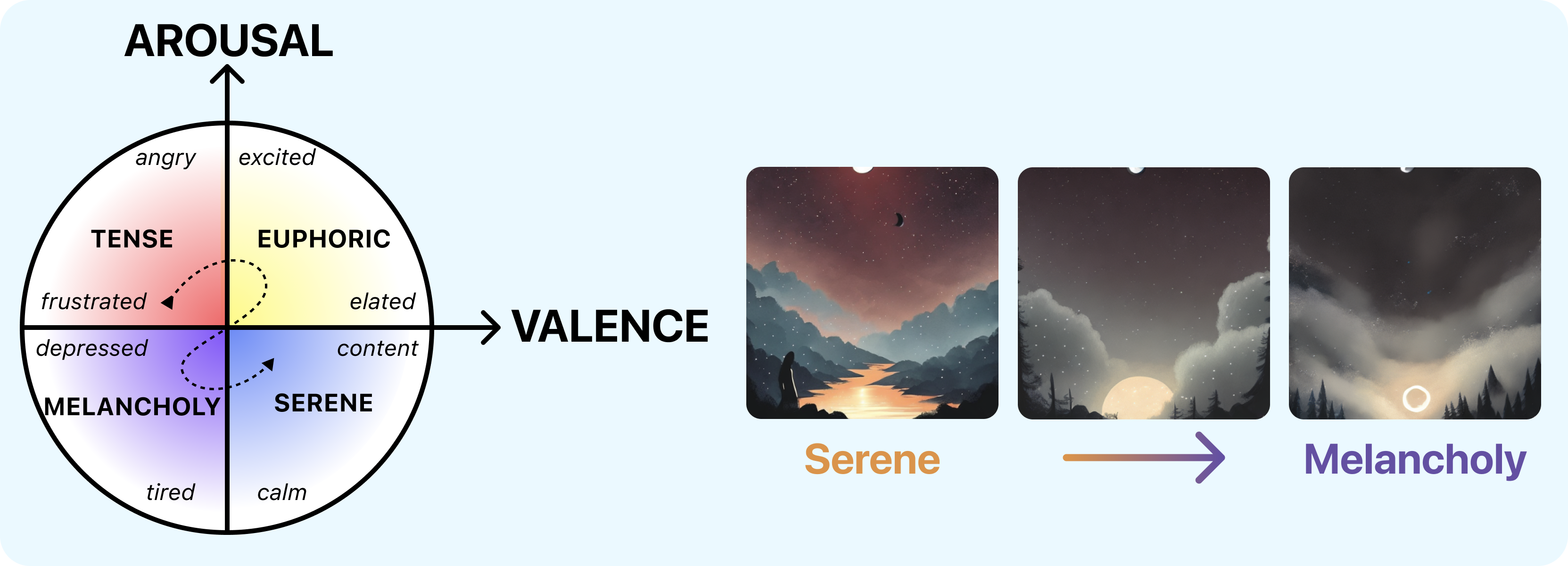}
    \caption{Left: valence/arousal emotion spectrum. Right: \textbf{serene}-\textbf{melancholy} spherical interpolation}
    \label{emo}
\end{figure}

\textbf{On-set strength guided spherical interpolation:} We craft seamless visual transitions between images using Giffusion's spherical interpolation \cite{giffusion} which synchronizes visual changes with emotional and lyrical shifts. For audio-reactive animation, we analyze the music's onset strength envelope to capture rhythmic elements. We change the interpolation weight scheduling to enable rapid transitions during beat-heavy segments and gradual evolution during quieter passages, resulting in a cohesive and rhythmically synchronized video.

\textbf{Stylization using finetuned checkpoint models:} Fine-tuning Stable Diffusion for specific styles is computationally intensive. Given our resource constraints, we leverage pre-existing style customization models from Civitai \cite{Civitai} finetuned on specific art/photos to capture the different styles like sketch: \cite{sketch}, western animation: \cite{west} and hyper realistic: \cite{real_vis}.

\subsection{Security and Personalization}

\textbf{Custom characters using LoRA:} We leverage Low-Rank Adaptation (LoRA) training to create personalized music videos, fine-tuning our image generation model with minimal user input. Using the kohya SS GUI \cite{kohya} and DreamBooth methodology \cite{ruiz2023dreambooth}, we train a custom LoRA adapter on Stable Diffusion \cite{stablediff} using 7 webcam-captured user images. This efficient and flexible process allows the trained LoRA adapter to work across multiple checkpoint models.

\begin{figure}[h!]
\centering
\includegraphics[width=\textwidth]{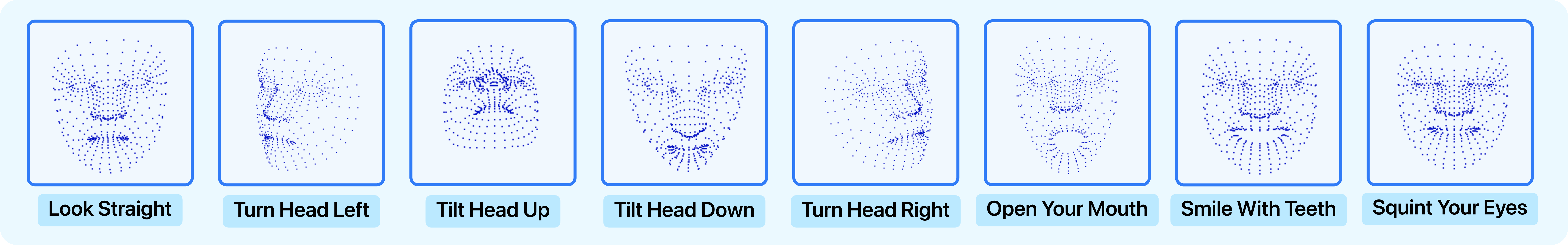}
\caption{7 CHARCHA Protocol Actions \& their backend detection using MediaPipe}
\label{charcha}
\end{figure}

\textbf{CHARCHA protocol:} The images for LoRA are obtained through our CHARCHA protocol which stands for: Computer Human Assessment for Recreating Characters with Human Actions. CHARCHA verifies the user's identity by prompting real-time actions which include a variety of facial expressions and head poses (\ref{charcha}) to ensure diversity. This approach enables seamless integration of the user's appearance in the generated music video. At the same time, it allows us to put safety first in generative AI applications online, ensuring that users cannot use other people's likeness to generate harmful or disturbing content. 


\textbf{Description of Protocol:}
Like CAPTCHA, our protocol is designed to be simple and fast (60-90 seconds). 

\begin{enumerate}
    \item \textbf{Calibration}: Each test begins with a calibration step that determines the position of a person's head in front of the camera. A image of the person with their head in neutral position is captured. This takes no more than 2 seconds. 
    \item \textbf{Action Sequence}: Following the calibration, users are asked to perform an action picked at random from the set of actions (\ref{charcha}) for 10 seconds. We record an image from the user when an action is successfully executed. This process is repeated for 6 actions (without duplicates) with a 5 second gap for users to prepare before the next action. 
\end{enumerate}

While the person is performing the actions, each movement performed by the user is verified for performing the action displayed to them in real-time. A score is recorded for each action out of 10. To pass, a person must successfully complete all six actions with a score of 6/10. This number is determined experimentally as shown in the results. If the person fails, they can retake the test once more. If this fails, CHARCHA verification is considered to have failed and the face cannot be used for generation. 

Unlike other generative AI tools like "Imagine Me" from Meta AI \cite{imagineme}, which captures 3 selfies of a person from different angles, the seven screenshots capture more diverse features of a person like their smile or stance while standing and help us generate expressive videos. As we show in the results, this allows us to create videos with greater resemblance to the user and more diversity in generation process. 

\section{Results}
\subsection{Music to Video} We strongly urge readers to view the generated videos in the \href{https://drive.google.com/drive/folders/1p7XzfnypVpkv_NGqIMurK5HF8cIUZLkF?usp=sharing}{drive folder} for results.

Checkpoint models significantly influence generation style (Figure \ref{personal_compare}), demonstrating the importance of model selection for desired aesthetic with a custom character. We compared three models using prompts from Rick Astley's ``Never Gonna Give You Up":

\textbf{Queratogray sketch \cite{sketch}:}  grayscale/monochrome illustrations creating a unique sketch-style aesthetic.

\textbf{Western Animation Diffusion \cite{west}:} animated aesthetic reminiscent of western animation, cartoon screencaps or comicbooks

\textbf{Realistic Vision model \cite{real_vis}: }Hyper-realistic images with human-like faces.

All models effectively preserve text prompt details based on lyrics, indicating robust text-to-image alignment across styles even with different character LoRAs loaded.
\subsection{Security \& Personalization}
We combine the realistic vision model with our character LoRA to accurately reproduce individuals in our music videos. This combination integrates authentic user appearances into the generated content, as our emperically demonstrated \ref{personal_compare}.

We test CHARCHA using three experimental phases. You can try the experiment \href{https://charcha-test-414789f3aa3d.herokuapp.com/}{here}.  

\textbf{Phase 1}: We ask participants to pick any one action from the list of nine actions and perform it randomly for 10 seconds. We do this to determine whether participants are comfortable doing the actions and come up with better thresholds for scoring and time constraints for the test. 

\textbf{Phase 2}: This is the actual CHARCHA test where participants perform a specific action for 10 seconds. This is repeated 6 times and their actions are verified using the mediapipe library \cite{mediapipe}.  

\textbf{Phase 3}: In this phase, we ask participants to get creative and break CHARCHA. Some examples of what the participants tried include: trying to perform several different actions at once, wear sunglasses or head coverings, occlude the webcam or use someone else's images in front of the webcam to try and pass the test. 
    
\begin{figure}[h!]
\centering
\includegraphics[width=\textwidth]{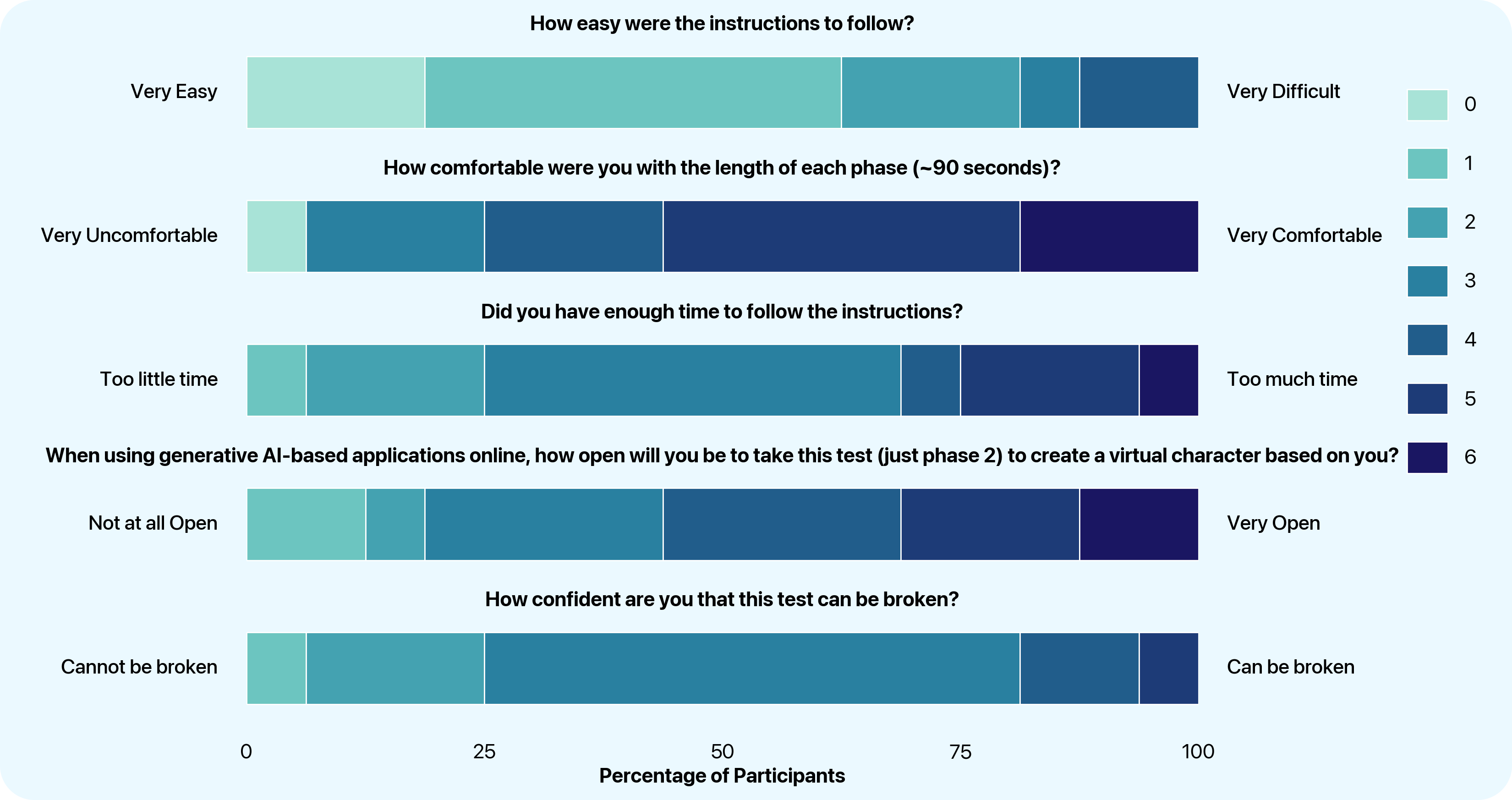}
\caption{Survey of CHARCHA experiment with n=16 participants}
\label{survey_results}
\end{figure}

The results of the survey conducted after the test are summarized in \ref{survey_results}.

\textbf{Measuring Reproducability of characters:} We evaluated how accurately AI-generated videos could reproduce participants' appearances from CHARCHA protocol images. Our process involved training a Dreambooth LoRA model \cite{hu2021lora,dreambooth} using CHARCHA user images. We then generated videos with the Stable Diffusion 1.5 Realistic Vision model \cite{real_vis} using the same prompts but different character LoRAs. To analyze reproducibility, we focused on 7 participants, comparing video frames to original CHARCHA images. The comparison utilized OpenCV \cite{opencv} for face detection and VGG-Face \cite{vggface} for face verification (which shows competitive accuracy using the DeepFace library \cite{deepface1, deepface2}). This method allowed quantitative assessment of AI's ability to recreate participant likenesses in generated videos (we average the results for all 7 generated videos):
\begin{table}[h]
    \centering
    \begin{tabular}{|c|c|c|}
        \hline
        \% of frames with participant's face & \% of frames with no face &  \% of face frames with participants face\\ \hline
        81\% & 11.1\% & 92\% \\ \hline
    \end{tabular}
    \vspace{0.2cm}  
    \caption{Face verification metric}
    \label{tab:face_verified}
\end{table}

Here,
\% of face frames with participants face = $\frac{\text{no.\,of\,frames\,with\,participants\,face}}{\text{total\,frames\,-\,no.\,of\,frames\,with\,no\,face}}$ $\times$ 100
\label{eq:total_verified_face_frames}

We compared CLIP image similarity scores between CHARCHA images and video frames in the appendix to verify that LoRA training didn't simply replicate training data. The observed score fluctuations throughout the video support our qualitative findings of diverse character settings, as seen in \ref{personal_compare}.

\section{Limitations \& Future Work}
As a novel formulation, this project has numerous areas for improvement. Current limitations include single-character personalization, inter-frame consistency, frame flickering, and prompt adherence. Future enhancements will focus on adopting multi-subject tuning-free personalization \cite{imagineme}, Stable Diffusion XL \cite{SDXL} for higher quality images and text-to-video models for improved coherence, potentially incorporating camera movements. We aim to strengthen CHARCHA, our facial verification protocol, against deep fake technologies like Deep Face Live \cite{deepfacelive}, expanding verification methods and defenses against impersonation. Additionally, we plan to implement hyper-parameter optimization for personalization and develop additional evaluation metrics for both personalization and music video generation.

\section{Conclusion}
In this paper, we introduced MVP, a fully-automated pipeline for personalized music video generation that integrates various multimodal translation and generation techniques. By leveraging advanced models in audio transcription, language modeling, text-to-image diffusion, and music emotion recognition, MVP provides a novel approach to creating music videos that are both personalized and immersive. The incorporation of LoRA for personalizing the visual content and the introduction of CHARCHA for secure facial identity verification highlight the innovative and ethical considerations at the core of this project.

\begin{ack}
We extend our gratitude to Professor Louis Philippe Morency for his guidance in conceptualizing Music Video Generation and emphasizing the integration of multimodal knowledge into the model. We also thank CHARCHA participants and face models (who have chosen to remain anonymous) for their invaluable contributions to enhancing the system.
 \end{ack}

\bibliographystyle{IEEEtran}
\bibliography{neurips_2024}

\newpage
\appendix

\section{Appendix / supplemental material}
\subsection{MVP Model Architecture}
\begin{figure}[h!]
\begin{center}
\includegraphics[width=0.8\textwidth]{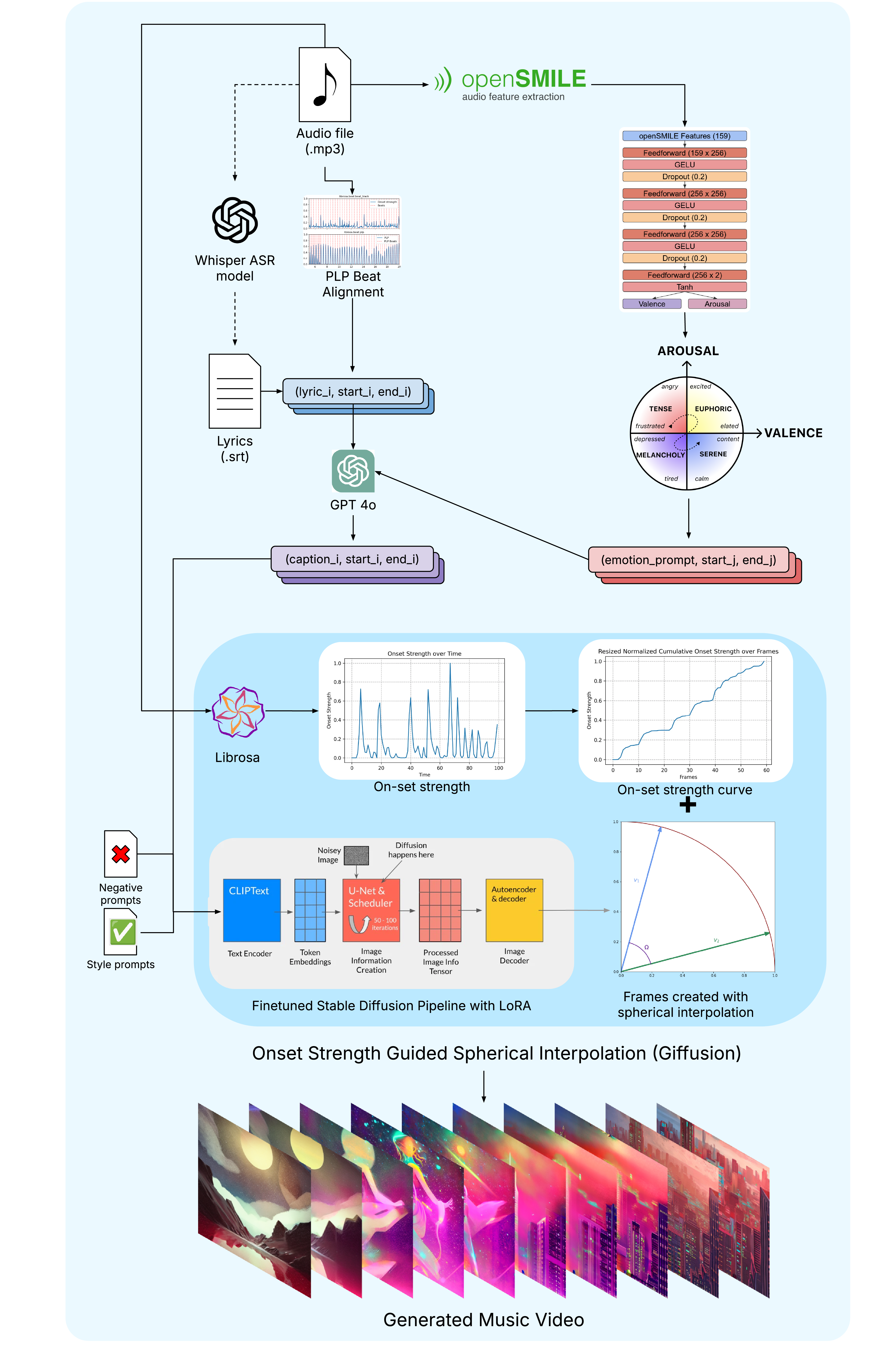}
\caption{MVP Model Architecture}
\label{model_architecture}
\end{center}
\end{figure}

Our application simply runs on one input, an MP3 file of the music. This is then broken down into multiple modalities for the purpose of generating a music video. Our application works as per the model diagram \ref{model_architecture}. Style of the video is changed by loading a different fine-tuned Stable Diffusion checkpoint. The CHARCHA images are used to tune the loaded LoRA adapter to generate specific characters in the Latent Diffusion model.

\subsection{Stylization through Checkpoint Models}
\begin{figure}[h!]
\includegraphics[width=\textwidth]{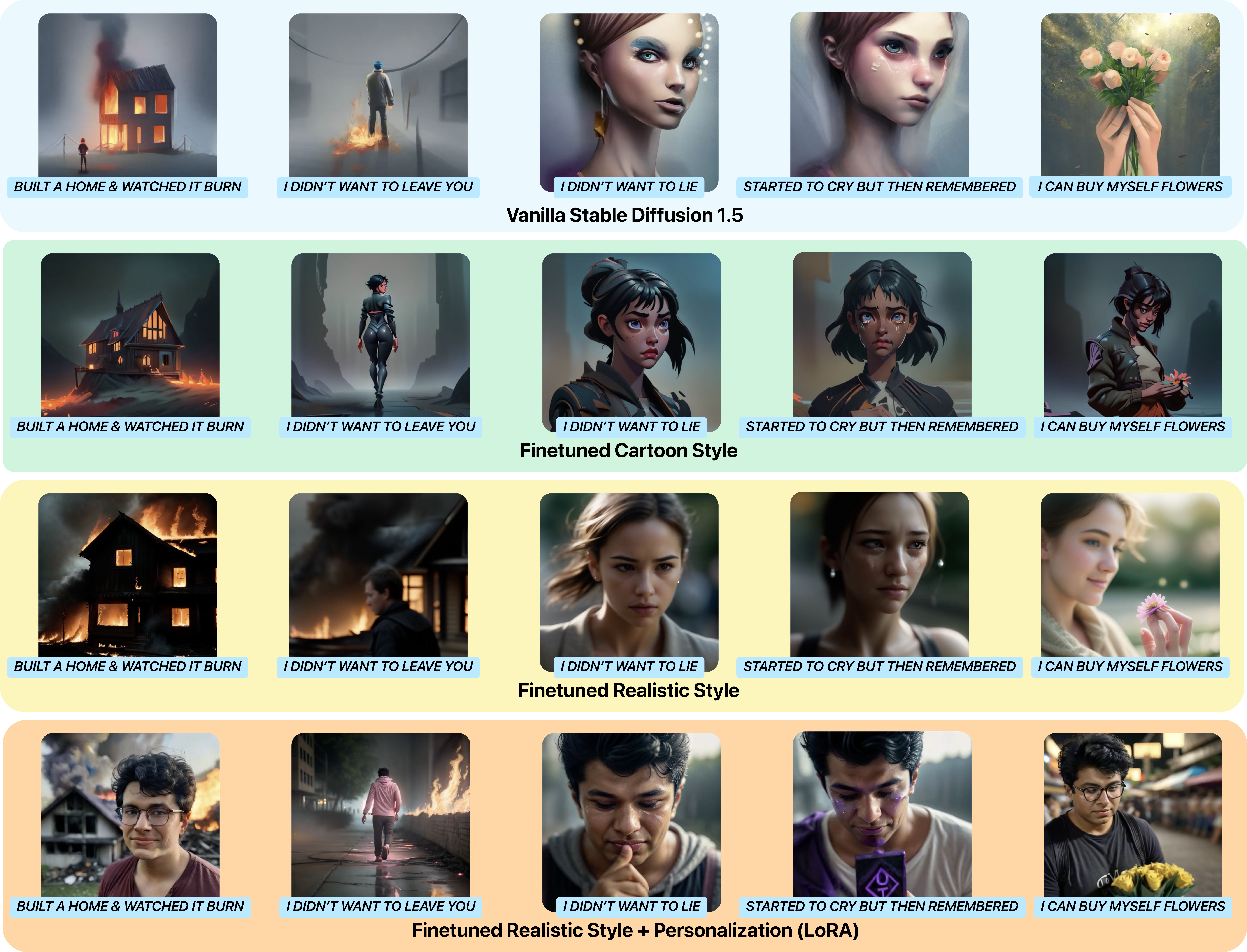}
\caption{Lyric and videoframe pairs using \textbf{vanilla SD 1.5}, \textbf{ToonYou Beta 6 checkpoint \cite{toonyou}}, \textbf{Realistic Vision V5.1 checkpoint \cite{real_vis}} and \textbf{Realistic Vision + trained character LoRA} in vertical order}
\label{flowers_grids}
\centering
\end{figure}

We display image stills from generated music video of the song "Flowers" by Miley Cyrus \ref{flowers_grids}. There is significant difference from baseline with the checkpoint models along with good adherence to the lyric. Additional style prompts (``cartoon" or ``realistic") with vanilla Stable Diffusion 1.5 proved unsuccessful (analysis omitted for brevity). Thus, our qualitative evaluation supports the consensus that style consistency is strongly dependent on checkpoint models rather than simple prompt engineering \cite{lia_paper, dreambooth}.

\begin{figure}[h!]
\centering
\includegraphics[width=\textwidth]{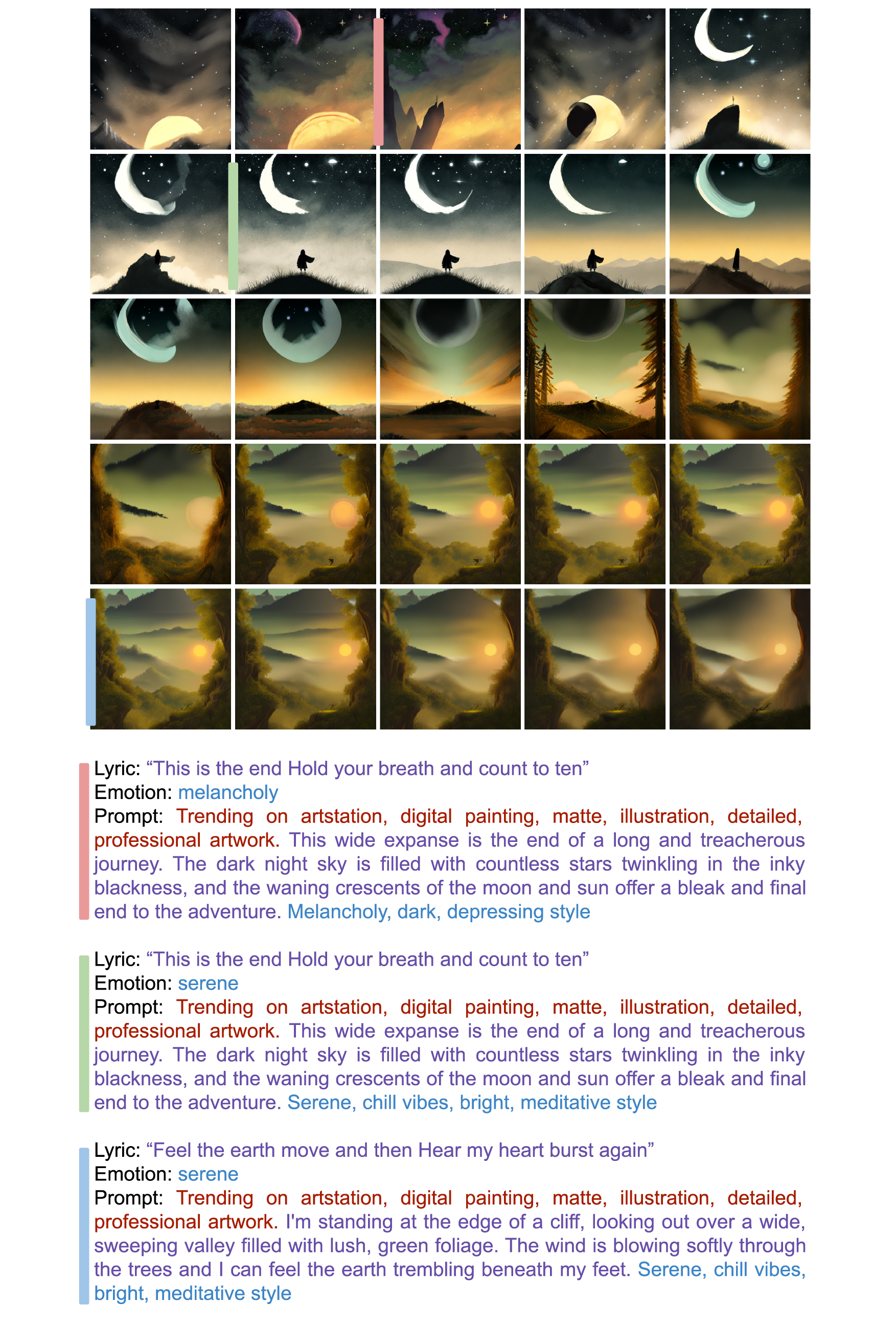}
\caption{A series of generated frames and corresponding prompts based on an excerpt of "Skyfall" by Adele. The color bars indicate when each prompt occurs.}
\end{figure}

\subsection{Evaluation of CHARCHA Images and Generated Video Frames}

For the same song (``Never Gonna Give You Up'' by Rick Astley), we use identical prompts to generate the music video, only changing the loaded Character LoRA trained on an individual's CHARCHA captured images. We use the Realistic Vision checkpoint model to aim for photo-realism.

\begin{figure}[h!]
\includegraphics[width=\textwidth]{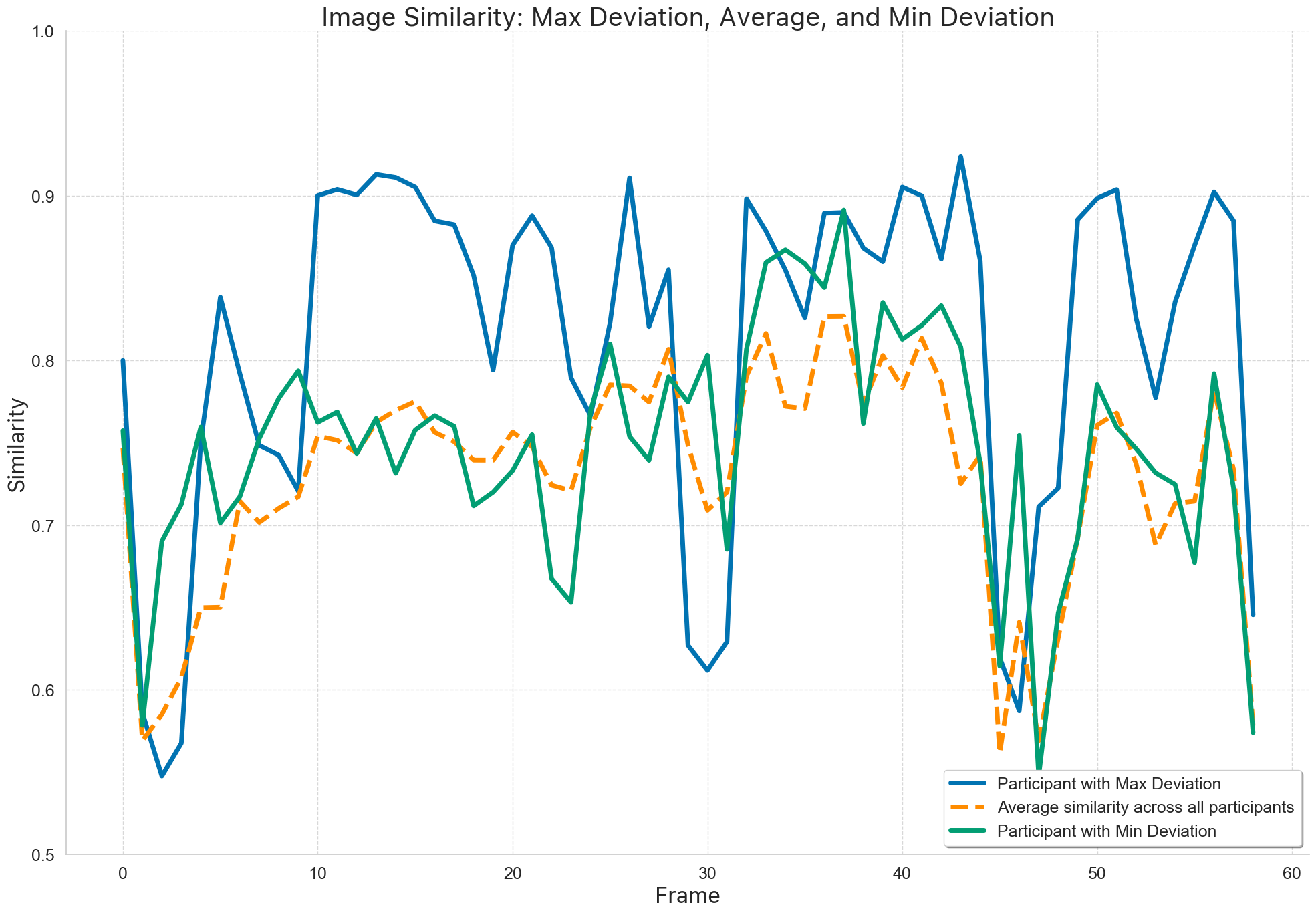}
\caption{CLIP Similarity between CHARCHA images and generated video frames}, 
\label{similarity_graph}
\centering
\end{figure}

\textbf{Similarity Analysis}: For each participant, we calculated the CLIP image-to-image similarity \cite{clipscore} between the reference images and the generated video frames for every second of a minute of video. We averaged the scores across each CHARCHA image captured for that individual (termed the character similarity score). In Figure~\ref{similarity_graph}, the average similarity represents the mean of 7 different character similarity scores at each timestep (averaged over 7 individuals). We also show the character similarity scores of the participants with maximum and minimum deviation.

\textbf{Interpretation of Results:} The dynamic nature of this graph, with its fluctuations, indicates that the video doesn't simply display images from the training data (the CHARCHA images). If that were the case, the graph would be very close to 1 and not exhibit such variability. However, there is evidence that the character is faithfully represented in the generated video, as the character similarity score ranges from 0.6 to 0.9. This is notably higher than what would be expected when comparing the CHARCHA images to a video generated using a different character.

\textbf{Limitations:} It's important to note that CLIP is known to be highly sensitive, and existing literature suggests it may not align well with human judgment \cite{clipeval}. Due to these potential limitations, this evaluation is presented in the appendix. We aim to get better evaluation for future iterations.

\subsection{Emotion Extraction Details}
\begin{figure}[h!]
    \centering        \includegraphics[width=0.4\textwidth]{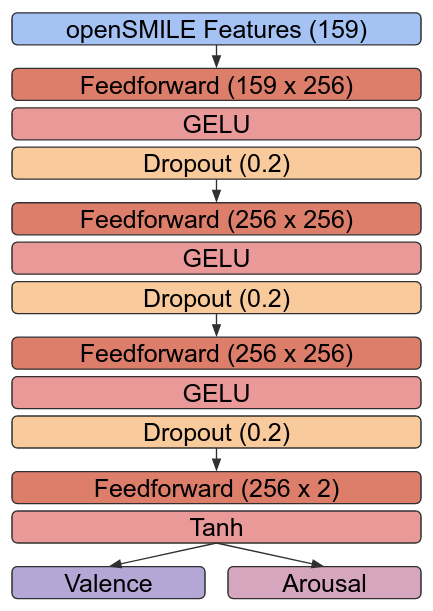} 
        \caption{Arousal/Valence Prediction Model}
    \label{va_model}
\end{figure}

In our framework \ref{va_model}, we extract openSMILE features \cite{opensmile} from the raw audio of the music, and train a neural network model on the DEAM dataset \cite{DEAM} to predict arousal and valence values from openSMILE features. We achieved a validation mean-squared-error of 0.0206. For dynamically changing emotion, we predict a unique arousal-valence pair for each 5-second window of the music and corresponding openSMILE features. We use these values to continually update a position in the arousal-valence space by keeping a running sum, and register a change in emotion whenever the position crosses into a new quadrant.

\subsection{Beat Alignment}
The lyric-timestamp pairs used to generate images in the music video are taken from Whisper \cite{whisper}, which may not be a 100\% accurate and lacks consideration for the music's rhythm and beats. To enhance this, we aim to align the images with the song's beats by extracting Predominant Local Pulse (PLP) information from the music using a sinusoidal kernel \cite{plp} using the librosa library \cite{librosa}. This alignment not only creates a more immersive experience but also conveys shifts in mood and emotion, making the music video more impactful. By harmonizing visuals with the music's rhythm, we establish a sense of continuity and flow throughout the video, enhancing viewer engagement and meaning.

\subsection{Ethical Considerations}
Our final pipeline on a few rare occasions generates explicit content, particularly nudity, often during quick transitions. This can be partially attributed to the presence of unsafe content in intermediate outputs from spherical interpolation, despite the initial prompts discouraging such content. Artist credit attribution for Stable Diffusion outputs is a complex issue, as we avoid referencing specific artists' unique styles in our prompts. Instead, we prioritize transparency regarding the technology used. Additionally, the model may exhibit harmful biases, including "whitewashing" and oversexualization effects. We have made efforts to mitigate these biases with prompt keywords such as "diversity" and negative prompts against harmful stereotypes. We acknowledge these limitations and advise some viewer discretion as we cannot exert complete control over the model's output. We aim to explore solving these issues in our future work.

\end{document}